\documentclass[letterpaper, 10 pt, conference]{ieeeconf}  

\IEEEoverridecommandlockouts                              

\usepackage{graphics} 
\usepackage{epsfig} 
\usepackage{mathptmx} 
\usepackage{times} 
\usepackage{amsmath} 
\usepackage{amssymb}  
\usepackage[ruled]{algorithm2e}
\usepackage{subfigure}
\usepackage{cite}
\usepackage{tabu}
\usepackage{booktabs}
\usepackage{multirow}
\usepackage{footnote}
\usepackage{threeparttable}
\usepackage{booktabs}
\usepackage{array}
\usepackage{threeparttable}
\usepackage[table]{xcolor}
\usepackage[letterpaper,top=60pt,bottom=43pt,left=48pt,right=48pt]{geometry}

\title{\LARGE \bf
	Peer-Assisted Robotic Learning: A Data-Driven Collaborative Learning Approach for Cloud Robotic Systems
}

\author{Boyi Liu$^{{1},{2}}$, Lujia Wang$^{1*}$, Xinquan Chen$^{1,3}$, Lexiong Huang$^{1,3}$ and Cheng-Zhong Xu$^{2}$
	\thanks{*This research is supported by the Shenzhen Science and Technology Innovation Commission (Grant Number JCYJ2017081853518789), the Guangdong Science and Technology Plan Guangdong-Hong Kong Cooperation Innovation Platform (Grant Number 2018B050502009) and the National Natural Science Foundation of China (Grant Number 61603376) awarded to Dr. Lujia Wang.}
	\thanks{$^{1}$Boyi liu, Lujia Wang (Corresponding Author), Xinquan Chen, Lexiong Huang are with Shenzhen Institutes of Advanced Technology (SIAT), Chinese Academy of Sciences. Boyi liu is with Joint Laboratory of
		Artificial Intelligence and Robotics of SIAT and University of Macau. {\tt\small by.liu@ieee.org}; {\tt\small lj.wang1@siat.ac.cn} {\tt\small xq.chen@siat.ac.cn}; 
	{\tt\small lx.huang@siat.ac.cn}}
	\thanks{$^{2}$Boyi Liu and Cheng-Zhong Xu are with the University of Macau.  {\tt\small czxu@um.edu.mo}}
	\thanks{$^{3}$Xinquan Chen and Lexiong Huang are also with the University of Chinese Academy of Sciences.}
}
\begin{document}
	\maketitle
	\thispagestyle{empty}
	\pagestyle{empty}

	\begin{abstract}
		A technological revolution is occurring in the field of robotics with the data-driven deep learning technology. However, building datasets for each local robot is laborious. Meanwhile, data islands between local robots make data unable to be utilized collaboratively. To address this issue, the work presents Peer-Assisted Robotic Learning (PARL) in robotics, which is inspired by the peer-assisted learning in cognitive psychology and pedagogy. PARL implements data collaboration with the framework of cloud robotic systems. Both data and models are shared by robots to the cloud after semantic computing and training locally. The cloud converges the data and performs augmentation, integration, and transferring. Finally, fine tune this larger shared dataset in the cloud to local robots. Furthermore, we propose the DAT Network (Data Augmentation and Transferring Network) to implement the data processing in PARL. DAT Network can realize the augmentation of data from multi-local robots. We conduct experiments on a simplified self-driving task for robots (cars). DAT Network has a significant improvement in the augmentation in self-driving scenarios. Along with this, the self-driving experimental results also demonstrate that PARL is capable of improving learning effects with data collaboration of local robots.
	\end{abstract}
	\section{INTRODUCTION}
	Enhanced by the powerful capacity of AI, robotics has achieved many complicated tasks that could not be accomplished before. Deep learning is applied in large robotic projects such as grasping \cite{1} and navigation \cite{2}. With more training samples, robots can cope with more kinds of scenarios. Owing to cloud computing availability, the concept of “cloud robotic system” \cite{3} is attracting more and more attention from the researchers nowadays.
	\begin{figure}[thpb]
		\centering
		\includegraphics[width=1\linewidth]{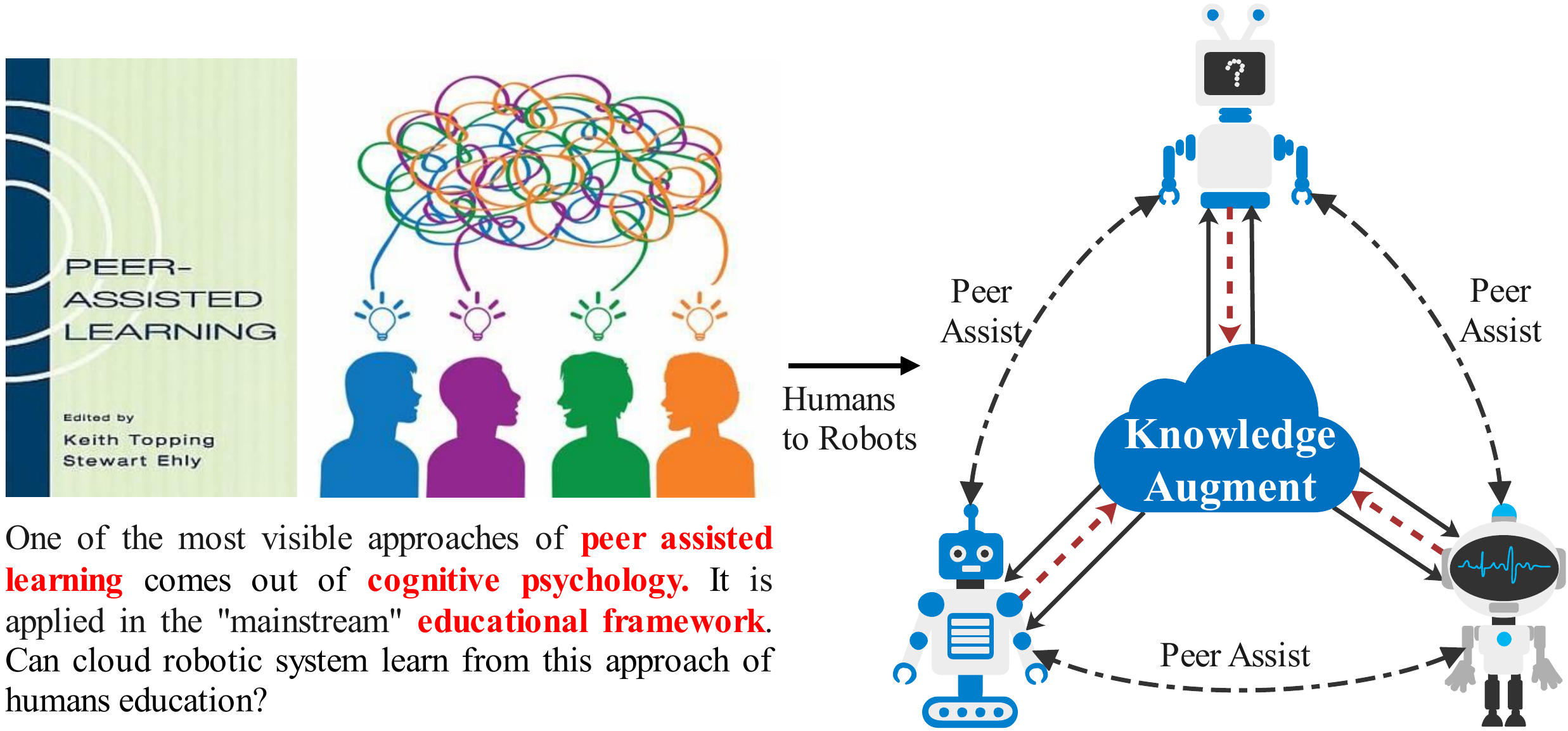}
		\caption{Peer-assisted learning has been researched in the field of cognitive psychology and pedagogy since the 1990s \cite{0}. Students can realize the goal of education by thinking collision and learning from each other. Inspired by this, we present Peer-Assisted Robotic Learning to solve data constraint and data islands, thus improving the learning ability. Robots realize peer assisting in cloud robotic systems with PARL.}
		\label{fig:architecture}
	\end{figure}
	Computing resources and information are obtained by robots through the cloud, which makes the intelligence of the robotic system expand to a broader scope.
	
	However, the high performance of learning methods usually demands numerous data and more computing resources. As computing can be offload to the cloud, data is gradually becoming the bottleneck restricting the robustness of robots. For instance, imitation learning needs lots of data as well as labels to cover the situations that robots may encounter. When data is insufficient, learning methods cannot generalize well to unseen scenarios, which eventually leads to mistakes. In reinforcement learning, robots need to interact with the environment. However, the acquisition of interaction is often expensive and time-consuming. Traditionally, the solution in computer vision is data augmentation, including random cropping \cite{4} , flipping \cite{5}, rotation, scale, and so on. The value distribution of pictures changes significantly in this way. But the distribution of objects remains unchanged. Therefore, augmentation can’t provide more semantic information for decision making. In automatic driving scenarios, the model trained with an insufficient dataset will cause deadly mistakes. It is worth seeking how to achieve data augmentation at the semantic level and enrich the content information, which facilitates decision making.
	
	In traditional machine learning, training data of the model need to be independent and identically distributed. Data from different distributions would not be processed together. On the other hand, data from various sources is gathered by the cloud robotic system and used by it. It is easier to excavate the value of data of different robots with its powerful computing capacity. For example, we can utilize the data collected in different cities and transfer them to the target domain in automatic driving, which significantly improves the object’s distribution in the training dataset.
	
    To avoid islands and enrich the datasets of local robots to enhance robotic systems' robustness, we propose the Peer-Assisted Robotic Learning (PARL) in the cloud robotic system. Contributions are as follows: 1. The work presents a novel framework with the implementation of PARL, which is capable of addressing data insufficiency and data island by peer-assisting. 2. We propose the DAT Network to augment the dataset of each self-driving robot in the cloud. It can realize a series of operations of data gathering, augmentation and transferring in PARL.
	\section{Related Theory}
	\subsection{Cloud Robotic System}
	James Kuffner first proposed the term “Cloud Robotics” \cite{6} in 2010. After that, more and more research on cloud robotics is emerging \cite{7}. With the development of big data and cloud computing, it is possible to design an effective, high real-time and low-cost cloud robotic system \cite{8}. The cloud robotic system's expectation is transferring lumbersome computing to the cloud and realizing the interaction of data between cloud and robots. A single robot doesn't need to equip with a powerful CPU. At the same time, various robots can share their data and technique with others\cite{LFRL}\cite{FIL}.
	
	Relied on cloud computing, many complex operations that require high real-time performance now can be achieved by a robot. For example, C2TAM \cite{9} is a cloud robotic framework designed for SLAM in robotics. Map optimization and storage are allocated to the cloud, and different robots can share their maps. A web community project named RoboEarth \cite{10} for robots to share knowledge was initiated by the Eindhoven University of Technology. Robots could easily acquire information and technique to meet their demands without going through complicated conversation process. \cite{12} suggested a resource allocation scheme for the cloud robotic system, which could improve resource utilization.
	Furthermore, cloud robotic technology also can be combined with the Internet of Things \cite{13}, optimizing the production process. The above research provided a broad imagination for cloud robotics. However, we need to be aware that many problems such as privacy and security concerns still exist in cloud robotics. Overall, cloud robotics is a promising research field.
	\subsection{Supervised Learning and Collaborative Learning on Robot}
	Supervised learning has been used to solve the training problem in robotics. Control policies often need to be provided as part of the input to the robot to learn. Foothold positions as well as features of the force-torque signal are supplied to the neural network \cite{14} to predict properties of the terrain ahead of the robot. The legged robot using this network can traverse diverse and rugged terrain. Through supervised learning,  \cite{15} cognitive robot can acquire corresponding concepts in human knowledge. Eventually, robots would perform better than humans in terms of relevance, accuracy and cohesiveness.
	
	Collaborative learning is one of the most critical topics in cloud robotics. It is about how robots collaborate with other robots to finish complex assignments \cite{16}. Sharing knowledge is an essential part of the collaboration, which means robots share knowledge, perception and technique with each other. \cite{17} presented a method to utilize OPENEASE cloud engine to exchange knowledge. By being connected, robots share the same structured knowledge and semantic rules. They can use other robots’ execution logs for reasoning. In \cite{18}, multiple robots can share their experience and learn a policy collectively. This work can shorten the data collection process and improve training time. Moreover, collaborative learning can be used in SLAM \cite{19}. Pose graphs obtained by multiple robots would be aggregated to generate a global pose graph. Because of its facility and practicability, supervised learning and collaborative learning are adopted in more and more robotic research.	
	\subsection{Peer-Assisted Learning}
	The process of PAL can be defined as “people from similar social groupings who are not professional teachers help each other to learn and learn themselves by teaching” \cite{20}. It’s a bidirectional process which means both sides are equal and beneficial. By means of PAL, teachers and learners are involved in the learning process greatly. Thus much more thorough comprehension is acquired by them. Nowadays PAL is mainly used in modern medical curricula \cite{21}. Related research \cite{22} shows that not only could PAL improve knowledge acquisition of peer teacher and peer learner, but also it could help them achieve better performance at the end of year the exam.
	
	Inspired by PAL, we present Peer-Assisted Robotic Learning (PARL). Robots can help each other and learn from other robots like PAL. Then, each robot's performance will be improved. In order to achieve this, we propose a new network model: Data Augmentation and Transferring Network (DAT Network). It can address the problem of data insufficiency in robotic learning and make robots perform better when encountering a new situation.
	\section{Methodology}
For this section, the work proposes the framework and procedure of PARL in cloud robotic systems in the first subsection. Then, the architecture of the DAT Network is proposed and described in the second subsection.
\subsection{Framework and Procedure of PARL}
	\begin{figure*}[thpb]
	\centering
	\includegraphics[width=1\linewidth]{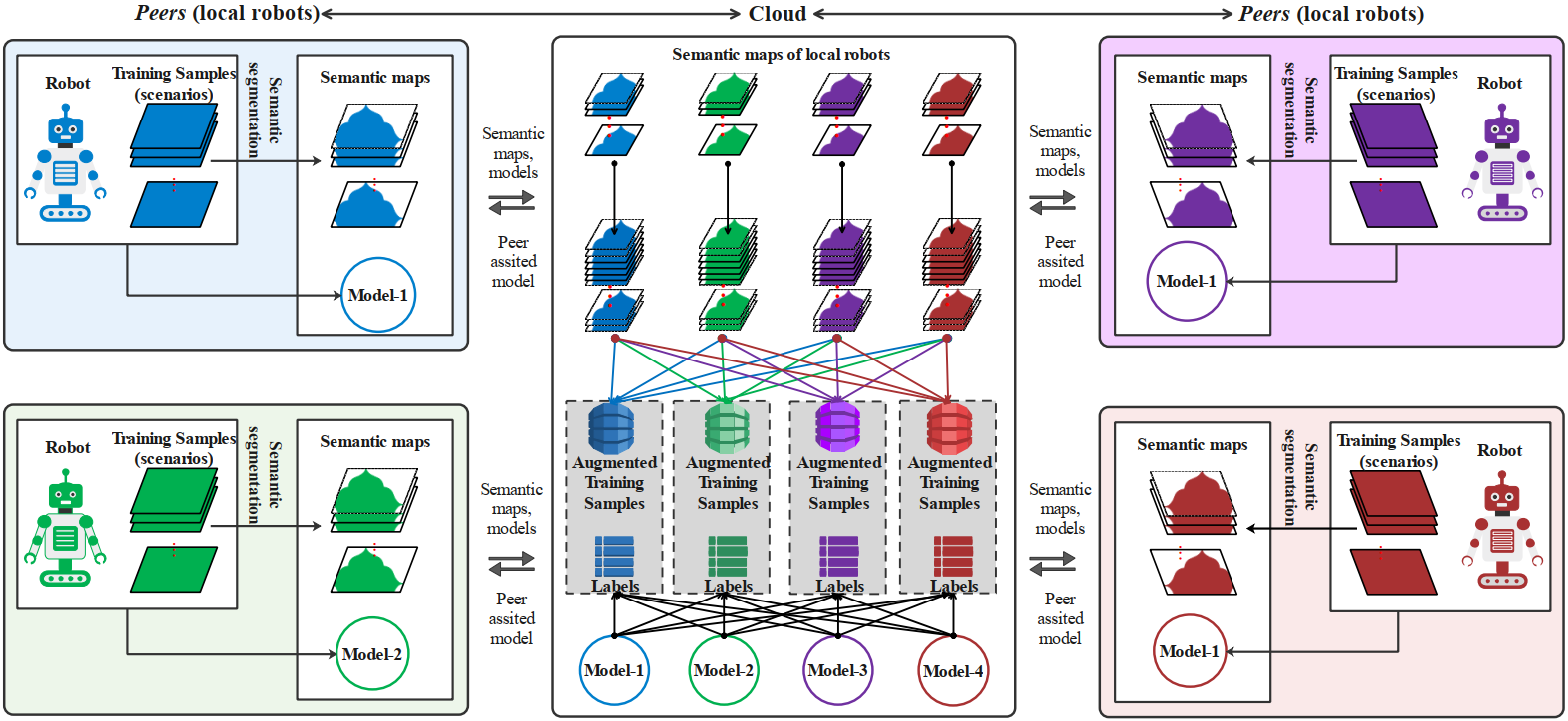}
	\caption{Framework and procedure of PARL. The box in the above figure filled with color represents the scenarios. The white and color interlaced boxes represent the semantic map. The arrows indicate the direction of data processing and transmission. The four-block diagrams on both sides respectively represent the calculation process of the local robot. The middle block diagram shows the data processing in the cloud, and the arrows of different colors represent different Pix2Pix models. The data of each robot is represented by a different color.}
	\label{fig:architecture}
\end{figure*}
Fig.2 presents the framework of PARL. The figure has consisted of five modules. The middle box represents data processing in the cloud. The four boxes around the middle box respectively represent the data processing of four local robots. We use $D_{l1}$, $D_{l2}$, $D_{l3}$ and $D_{l4}$ to represent the four local datasets corresponding to training samples of each local robot. $A_{l1}$, $A_{l2}$, $A_{l3}$ and $A_{l4}$ represent semantic datasets corresponding to segmentation images in the figure. $S_{a1}$, $S_{a2}$, $S_{a3}$, $S_{a4}$ represent the four augmented semantic datasets and they are presented as the second-row icons in the middle box. $C_1$, $C_2$, $C_3$, and $C_4$ represent four augmented datasets for four local robots. The corresponding labels of $C_1$, $C_2$, $C_3$ and $C_4$ are denoted by $label_1$-$label_4$.
\begin{figure}[thpb]
	\centering
	\includegraphics[width=1\linewidth]{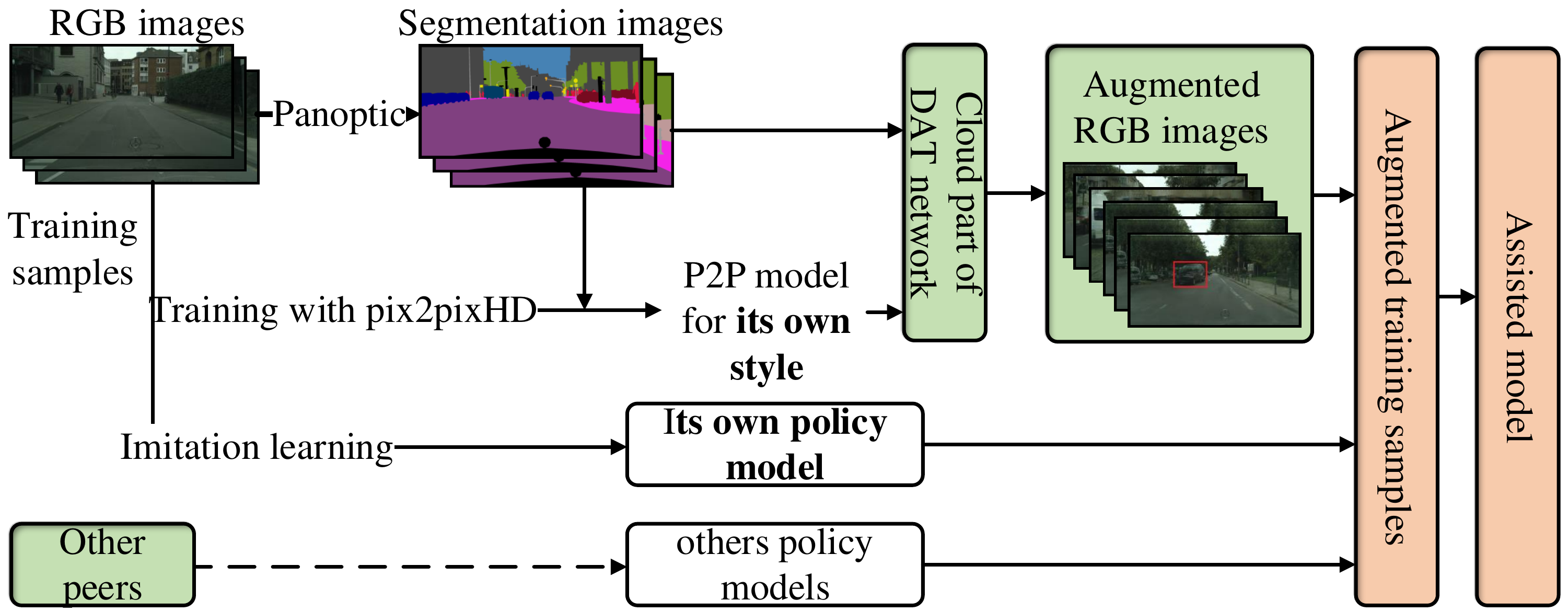}
	\caption{Example of data processing in one local robot. Black arrows denote the data direction. Colored boxes represent computing parts in the cloud or modules from other robots. Left images indicate local scenarios, middle images indicate semantic maps and right images indicate augmented scenarios.}
	\label{fig:architecture}
\end{figure}\\
\textbf{Computing in local robots.} The computing tasks of local robots are mainly semantics. The outputs of local robots are Pix2Pix model \cite{pix2pix}, semantic maps and instance maps. First, if there are no matching samples of scenarios and semantic maps. In order to acquire instance maps and semantic maps, the local robots need to perform semantic segmentation for the local dataset. Local robots train Pix2Pix model to transfer semantic maps to scenarios. Pix2PixHD\cite{ppHD} is an improved version of the Pix2Pix, so we apply Pix2PixHD actually in work. The trained Pix2PixHD networks in local robots will be transmitted to the cloud for data augmentation and style transferring in the cloud. The key is that the Pix2PixHD network of each local robot is trained separately. So, each local Pix2PixHD network possesses different parameters to translate semantic maps into scenarios. This corresponds to the second dimension of data augmentation in the cloud and the expansion of image style. After this step, the computing of the local robot is completed. Pix2PixHD model, semantic maps and instance maps will be transmitted to the cloud.\\
\begin{figure*}[thpb]
	\centering
	\includegraphics[width=0.8\linewidth]{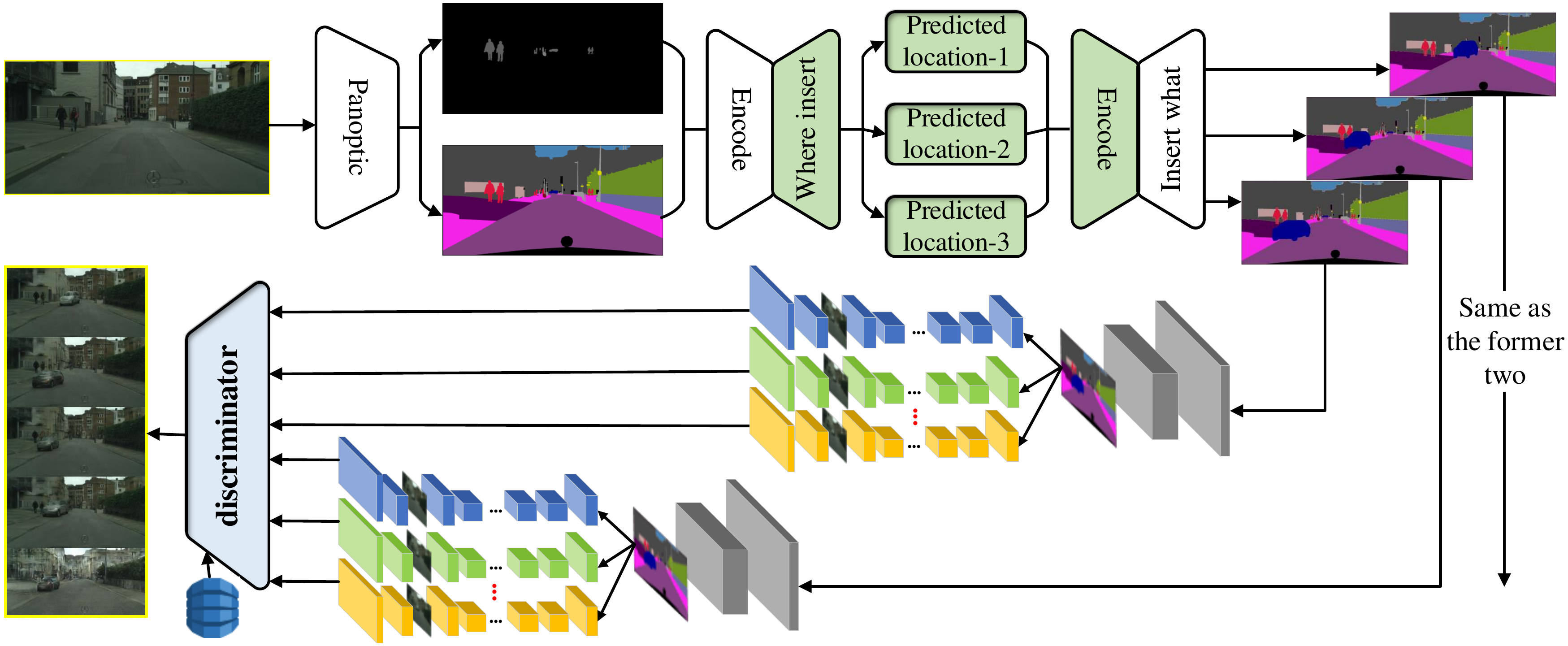}
	\caption{The framework of DAT Network. The input of the DAT Network is one scenario image and the output of the DAT Network is multi scenario images. A trapezoid placed vertically represents a neural network that includes the functions of an encoder or functions described on it. The arrows indicate the direction in which the data is transferred. The different colored network layers below represent different local robots, and the same colored ones represent the same parameters and structure. The blue icon in the left bottom denotes dataset of real scenarios.}
	\label{fig:architecture}
\end{figure*}
\textbf{Computing in the cloud.} Cloud augments the datasets that are transformed from local robots. The augmentation process is completed by the cloud part of the DAT network. This process can be understood as two stages: the augmentation of semantic maps and the augmentation through different Pix2Pix networks. So in the middlebox in the figure, the first step we show is to perform the augmentation for the basic semantic dataset of each local robot $A_{l1}$, $A_{l2}$, $A_{l3}$, $A_{l4}$ to $C_ 1$, $C_ 2$, $C_ 3$, $C_4$. Then, the augmented semantic maps assembly generates scenarios through different pix2pix networks to complete the second-dimension expansion. Since the generated data has no label, we label each data sample by crowdsourcing the local robot model. The shared model can be generated with the labeled data. Finally, the cloud model will be transferred to the local side to do fine tuning. Then we can get the decision model that can make use of all the data in the cloud robotic system.
	\subsection{DAT Network}
	As can be seen from the above, DAT Network is the key to realize PARL. In this section, we present a comprehensive DAT Network. It shows the complete steps of generating multiple scenarios from one scenario. These steps are interconnected and interact with each other, although DAT Network is composed of several stages. Therefore, the DAT Network should be regarded as a whole. It can be used as a function module to augment the self-driving data in the case of offline.
	
	As shown in Figure 4, it is the architecture of the DAT Network. A single image of the scenario is input into panoptic to obtain an instance map and a semantic map. Then, the scenario, instance map and semantic map are input into the “Where prediction” module and “What prediction” module \cite{ii} for semantic level data augmentation. For example in the figure, one semantic map is augmented to three. The “Where and What modules” are trained by combining the STN (Spatial Transformer Networks) and GAN (Generative Adversarial Network), which is inspired by the method from \cite{ii}. DAT Network input semantic maps to the Pix2Pix Network to generate scenarios. Pix2Pix networks have different styles because of different datasets of local robots. As shown in the figure, it has the same down-sampling layer, but the conversion layer and the up-sampling layer are different. These complete the second augmentation, image style augmentation. Different scenario styles are indicated in these different local Pix2Pix networks. Finally, a discriminator is added to determine whether the generated scenario is reasonable. The discriminator is trained by images generated and real scene datasets. The network retains the generated image when the output of the discriminator is higher than a threshold. Otherwise, the generated image is considered to be of low quality and it will be thrown away. The objective of the training of the DAT Network\cite{ii}\cite{ppHD}:
	\begin{equation}
		\left\{
		\begin{aligned}
		    \min _{G} \max _{D_{1}, D_{2}, D_{3}} \sum_{k=1,2,3} {L}_{\mathrm{G}}\left(G, D_{k}\right)\\
   			\min _{G_{il}} \max _{D{il}} {L}_{il}\left(G_{il}, D_{il}\right) \\
		    \min _{G{is}} \max _{D{is}} {L}_{is}\left(G_{is}, D_{is}\right)
		\end{aligned}
		\right.
	\end{equation}
	The above equation 1 respectively represent the three optimal conditions we need to obtain. The $L_{il}$ indicate the loss of insert layout network. The $G_{il}$ and $D_{il}$ respectively indicate the loss of generator and discriminator network of layout prediction. 
	\begin{equation}
		\begin{aligned}
		{L}_{il}(G_{il}, D_{il})={L}_{il}^{adv}(G_{il}, D_{layout}^{box})+{L}_{il}^{recon}(G_{il})
		+{L}_{il}^{sup}(G_{il}, D_{affine})
		\end{aligned}
	\end{equation}
	\begin{equation}
    	\begin{aligned}
        	{L}_{is}\left(G_{is}, D_{is}\right)={L}_{is}^{ad v}\left(G_{is}, D_{\text{layout}}^{\text {instance}}\right)+{L}_{is}^{\text {recon}}\left(G_{is}\right)+\\
        	{L}_{is}^{\text {sup}}\left(G_{is}, D_{\text{shape}}\right)
    	\end{aligned}
	\end{equation}
	$D_{\text{layout}}^{\text {box}}$ and $D_{\text{layout}}^{\text {instance}}$ focus on finding whether the new bounding box and instance fits into the layout of the input semantic map. $D_{affine}$ and $D_{shape}$ aim to distinguish whether the transformation parameters are realistic. $D_{il}$ and $D_{is}$ can denote the above discriminators that are related to the location prediction and shape prediction. Then, a mini-max game between $G_l$ and $D_l$ is formulated as above formula 2, 3. 
	\begin{equation}
		LFM\left(G, D_{k}\right)=\mathbb{E}_{(\mathbf{s}, \mathbf{x})} \sum_{i=1}^{T} \frac{1}{N_{i}}\left[\left\|D_{k}^{(i)}(\mathbf{s}, \mathbf{x})-D_{k}^{(i)}(\mathbf{s}, G(\mathbf{s}))\right\|_{1}\right]
	\end{equation}
	\begin{equation}
		\min _{G}\left(\left(\max _{D_{1}, D_{2}, D_{3}} \sum_{k=1,2,3} {L}_{\mathrm{G}}\left(G, D_{k}\right)\right)+\lambda \sum_{k=1,2,3} {L}_{\mathrm{FM}}\left(G, D_{k}\right)\right)
	\end{equation}
In the above formulas, D1, D2 and D3 are trained at the three different scales. the ith-layer feature extractor of discriminator $D_k$ as $D^{(i)}_k$. The feature matching loss LFM(G, $D_k$) is then as formula 4. T denotes the total number of layers, $N_i$ is the number of elements in each layer. Therefore, the new full objective update to the formula
5.
	\section{Experiments}
		\begin{figure*}[thpb]
		\centering
		\includegraphics[width=1\linewidth]{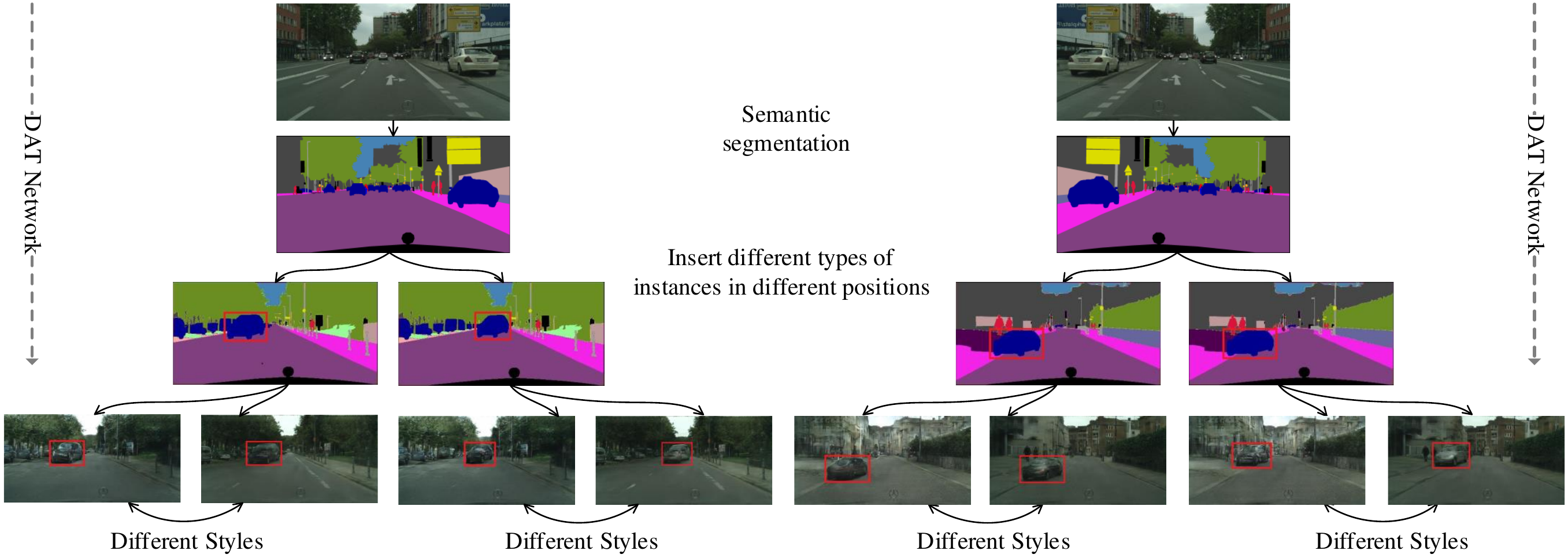}
		\caption{Image augmentation samples with DAT Network. DAT Network expanded the two scenarios to 8 scenarios. In this process, DAT Network augmented the scenarios in two dimensions, semantic and image style. We have adopted a 2-fold increase in each dimension. It can be seen from the figure that after the scenarios transformed into the semantic maps, the instances in it are added by DAT Network. Different pix2pix networks are used in the process of transferring segmentation images to scenarios. The image styles generated by each pix2pix network are different, so the data is also augmented in the image style dimension.}
		\label{fig:architecture}
	\end{figure*}
	\begin{table*}[htbp]
		\centering
		\caption{Comparison of augmented results by different methods}
		\begin{threeparttable}
			\begin{tabular}{crrrrr}
				\toprule
				Color jitter     &\includegraphics[width=0.15\linewidth]{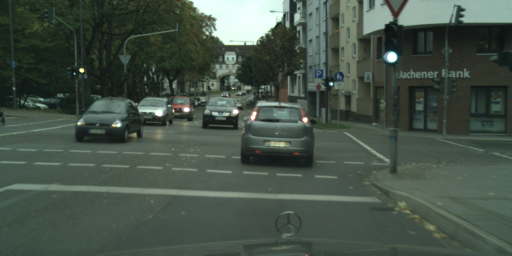}&\includegraphics[width=0.15\linewidth]{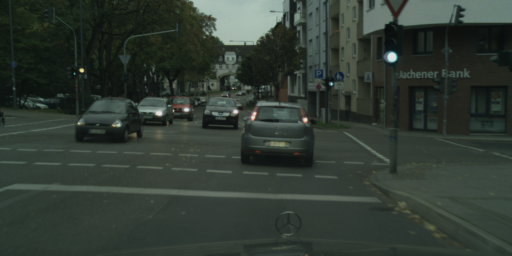}&\includegraphics[width=0.15\linewidth]{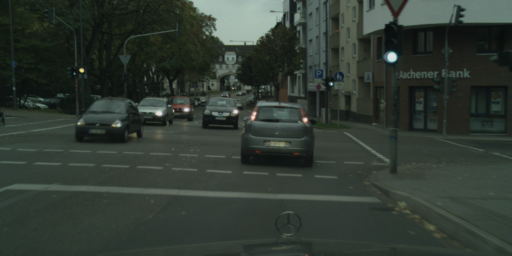}&\includegraphics[width=0.15\linewidth]{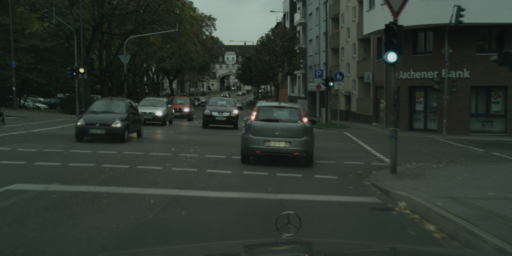}&\includegraphics[width=0.15\linewidth]{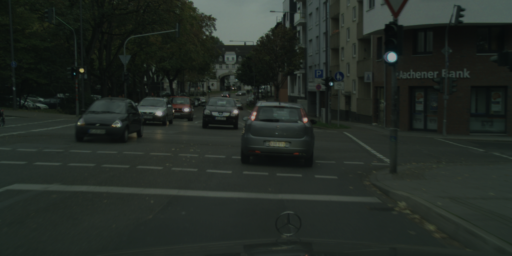}  \\
				\midrule
				Random resized crop &\includegraphics[width=0.15\linewidth]{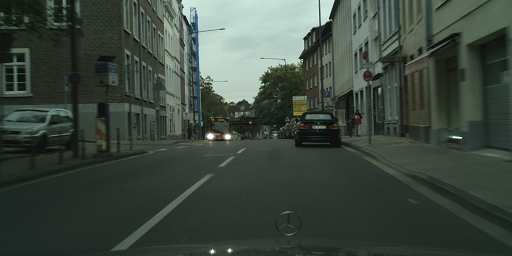}&\includegraphics[width=0.15\linewidth]{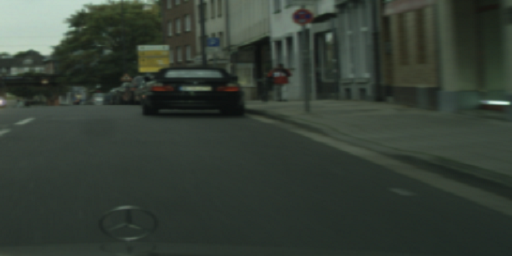}&\includegraphics[width=0.15\linewidth]{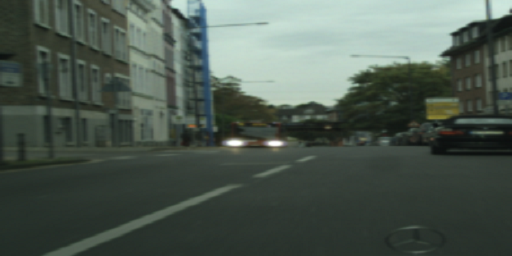}&\includegraphics[width=0.15\linewidth]{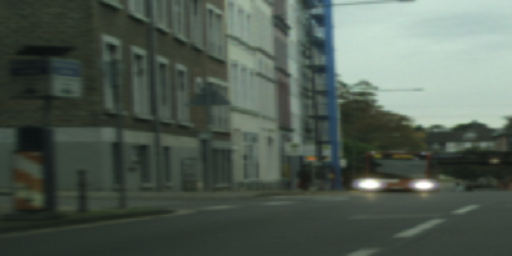}&\includegraphics[width=0.15\linewidth]{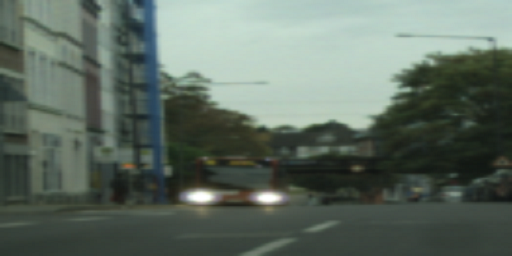}  \\
				\midrule
				\multirow{2}[4]{*}{DAT Network} &\includegraphics[width=0.15\linewidth]{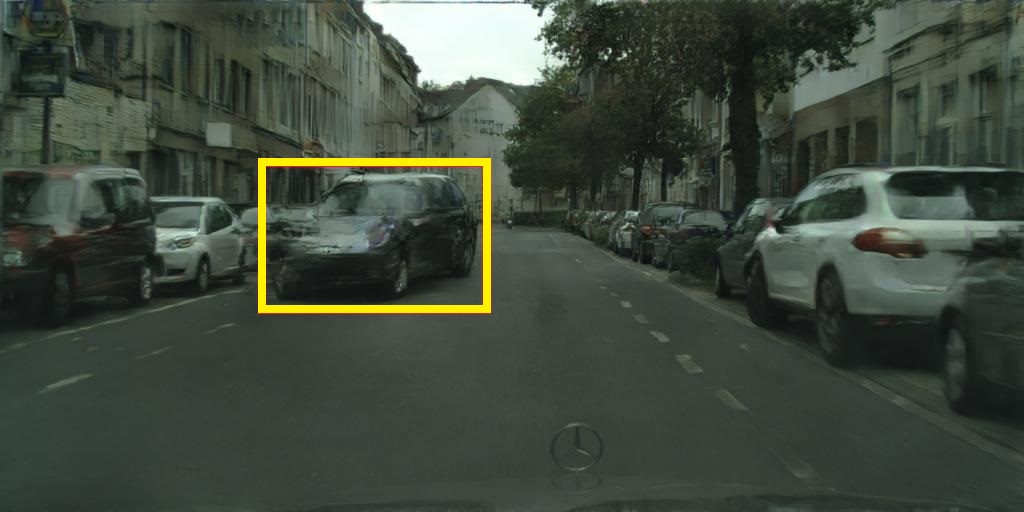}&\includegraphics[width=0.15\linewidth]{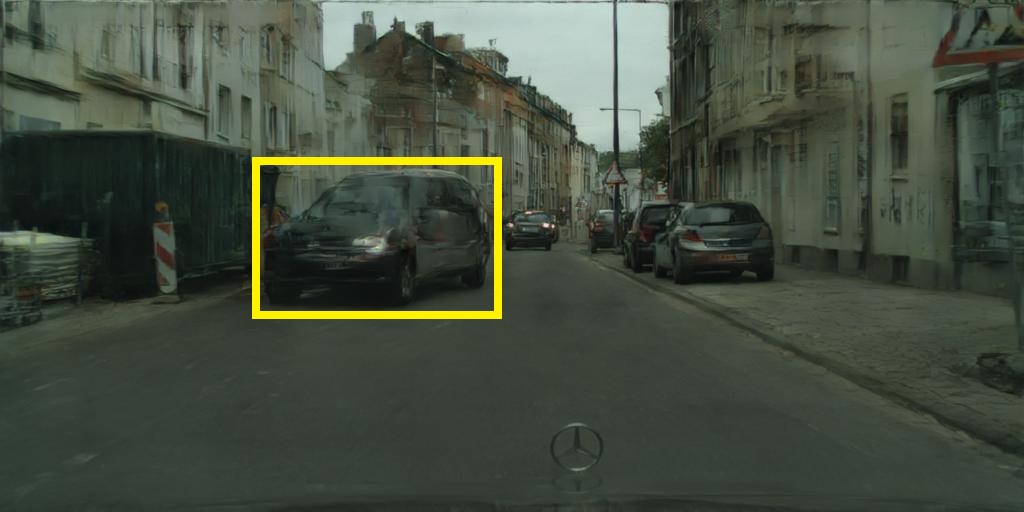}&\includegraphics[width=0.15\linewidth]{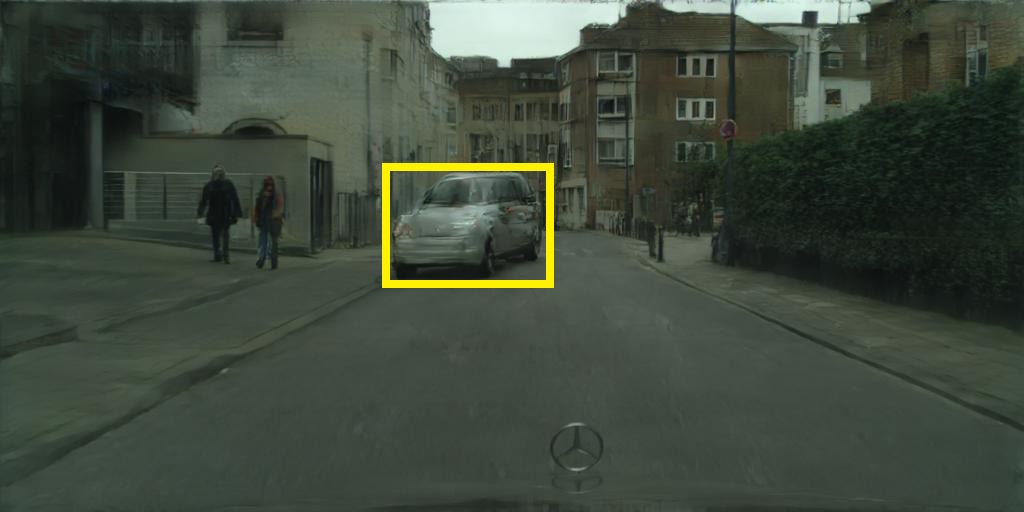}&\includegraphics[width=0.15\linewidth]{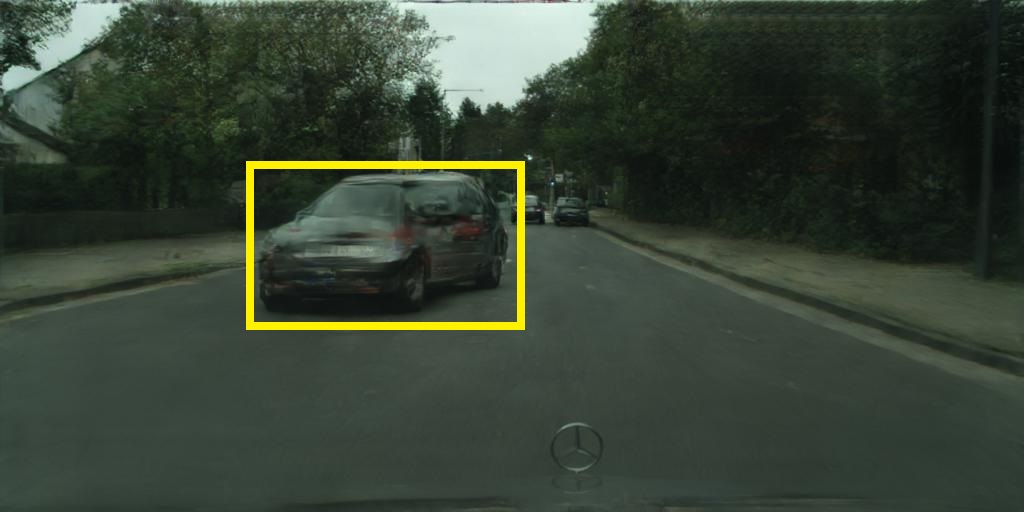}&\includegraphics[width=0.15\linewidth]{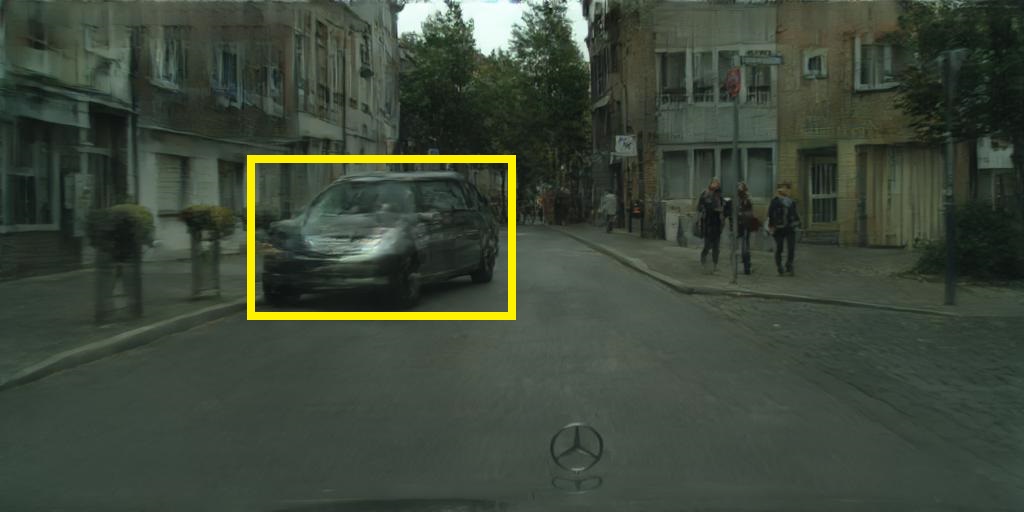}  \\
				\cmidrule{2-6}          &\includegraphics[width=0.15\linewidth]{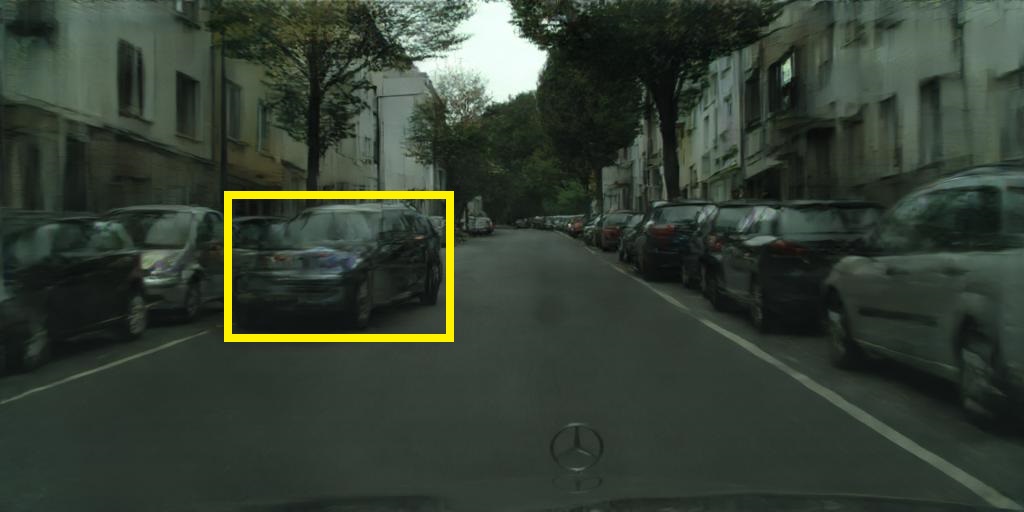}&\includegraphics[width=0.15\linewidth]{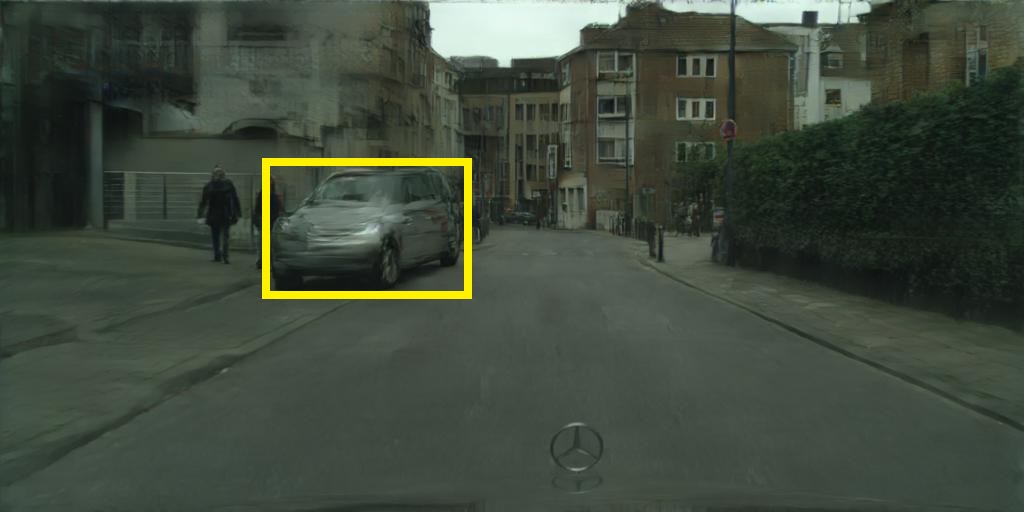}&\includegraphics[width=0.15\linewidth]{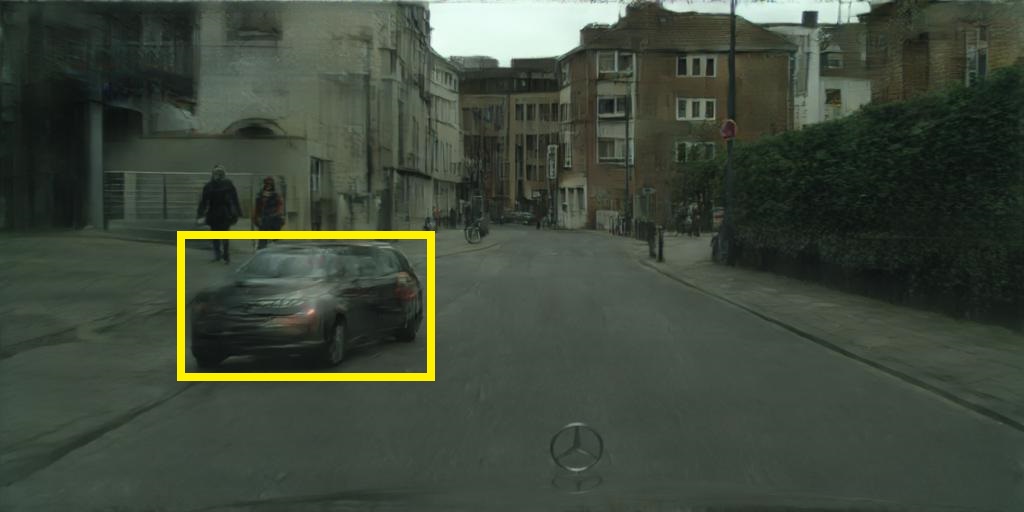}&\includegraphics[width=0.15\linewidth]{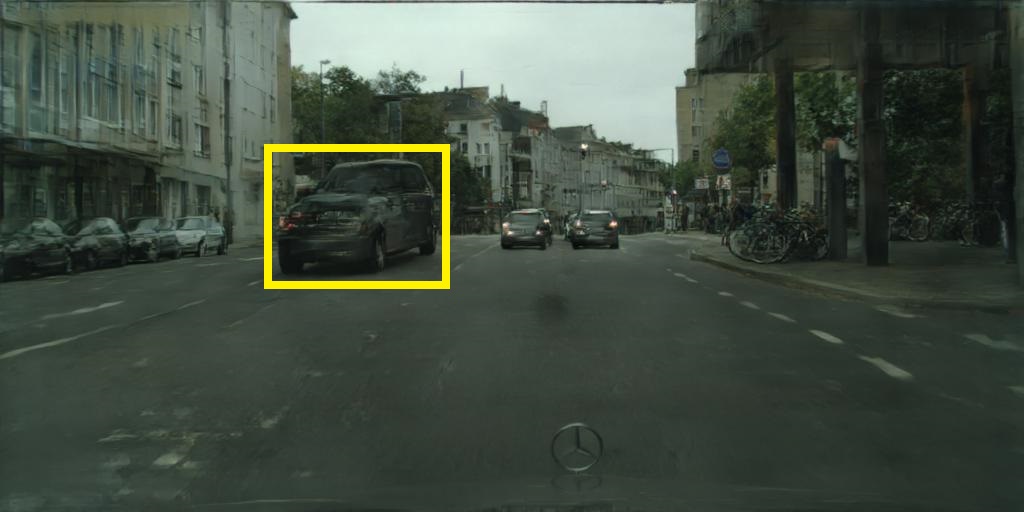}&\includegraphics[width=0.15\linewidth]{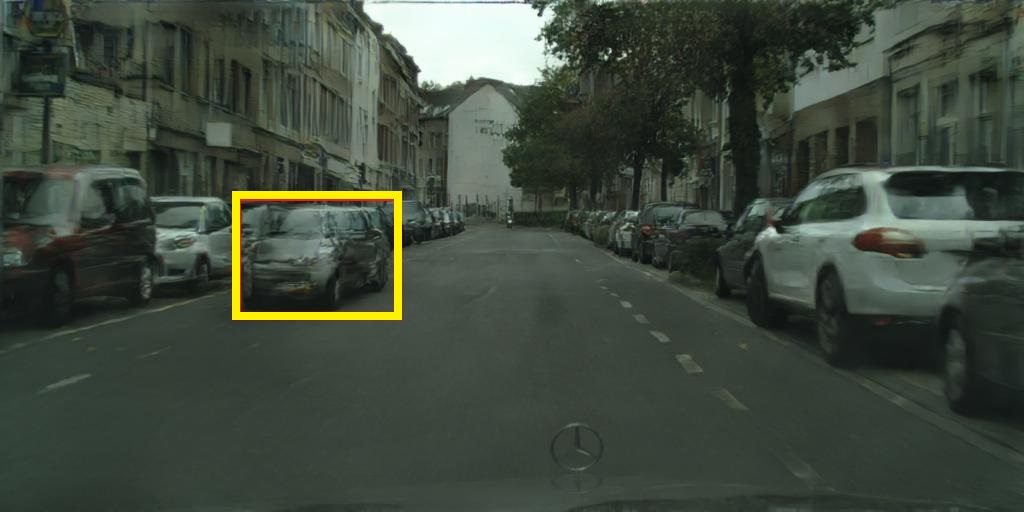}  \\
				\bottomrule
			\end{tabular}%
			\label{tab:addlabel}%
			\begin{tablenotes}    
				\footnotesize              
				\item[1]Color Jitter and Random Resized method augmented one scenario to five. DAT Network augmented three scenarios to ten. The yellow box is new instances that was generated.
				\item[2]As seen, DAT Network generates samples that are more similar to real scenarios than the other methods. The data augmentation of the other two methods is only geometric transformation, which has gap to real scenarios.
			\end{tablenotes}           
		\end{threeparttable}
	\end{table*}%
	In this section, we conduct experiments to verify the effectiveness of PARL on addressing data insufficiency in local robotic learning. The scene of our experiments is a self-driving case because a self-driving car is actually an advanced robot. Our experiments aim to answer the following two questions: (1) Assisted by other robots in the cloud robotic system, can DAT Network in PARL generate training data corresponding to the distribution of training samples in the original robot? (2) Under the condition of insufficient data, can PARL improve the local robot's learning effect in a cloud robotic system? For the first question, we split the self-driving data by several different locations in the Cityspace dataset\cite{Cordts2016Cityscapes} as different agents, and then carried out a data augmentation experiment on one agent based on DAT Network. To answer the second question, we conduct a self-driving experiment to compare the performance of the generic imitation learning and PARL. By performing robot executions of the two approaches, we can see that PARL is superior to imitation learning in many aspects.

	\subsection{Experiment of DAT Network in Peer-Assisted Learning}
	In this section, we conduct the experiment of the DAT Network to verify the effectiveness and compare it with other methods. The experimental details and results are shown as follows:\\
	\textbf{Experimental setup.} In the work, we used Cityscapes Dataset. The Cityscapes dataset comprises a large, diverse set of stereo video sequences recorded in streets from 50 different cities. Five thousand of these images have high-quality pixel-level annotations; 20000 additional images have coarse annotations to enable methods that leverage large volumes of weakly-labeled data. In the experiment of DAT Network in Peer-Assisted Learning, we divide the dataset into different parts based on the style of scenarios. Therefore, a different part of the dataset has a different distribution. Each part represents a different local robot in the cloud robotic system. Local robots acquire skills by training their own part of data in general imitation learning. DAT Network needs to implement data augmentation corresponding to the data distribution of local robots.\\
	\textbf{Baselines.} We compare to the following methods:\\
    - The color jitter method. Data enhancement of image color: changes of image brightness, saturation, contrast and other related factors.\\
    	    \begin{table*}[thbp]
    	\centering
    	\caption{Performance of controllers based on PARL and generic imitation learning in key challenging tasks}
    	\begin{threeparttable}
    		\begin{tabular}{cccc}
    			\toprule
    			\rowcolor[rgb]{ .851,  .851,  .851} \multicolumn{4}{c}{Turns} \\
    			\midrule
    			\includegraphics[width=0.20\linewidth]{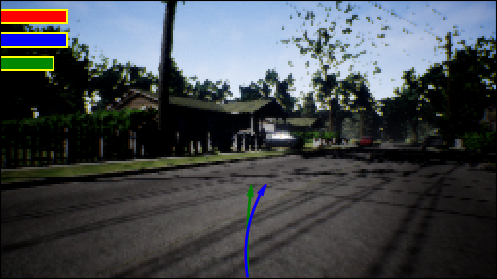}     & \includegraphics[width=0.20\linewidth]{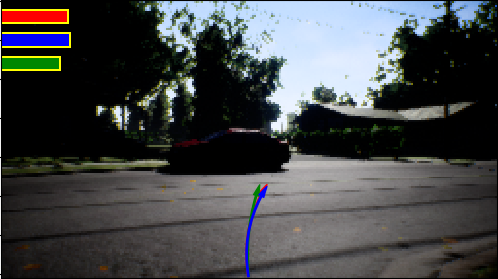}     & \includegraphics[width=0.20\linewidth]{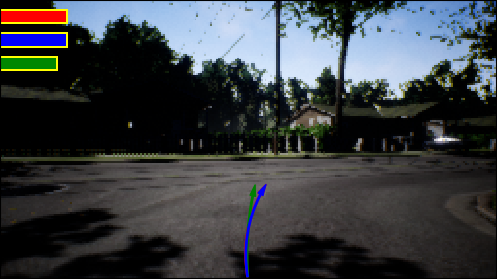}     & \includegraphics[width=0.20\linewidth]{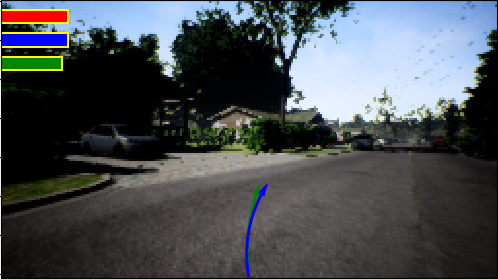} \\
    			E-PARL:0.00, E-IL:0.11 & E-PARL:0.02, E-IL:0.06 & E-PARL:0.00, E-IL:0.06 & E-PARL:0.00, E-IL:0.03 \\
    			\midrule
    			\rowcolor[rgb]{ .851,  .851,  .851} \multicolumn{4}{c}{Avoid cars} \\
    			\midrule
    			\includegraphics[width=0.20\linewidth]{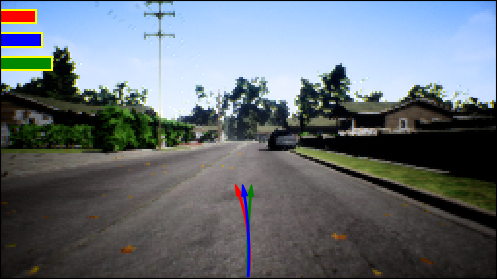}     & \includegraphics[width=0.20\linewidth]{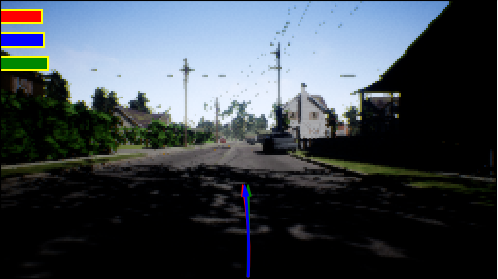}     & \includegraphics[width=0.20\linewidth]{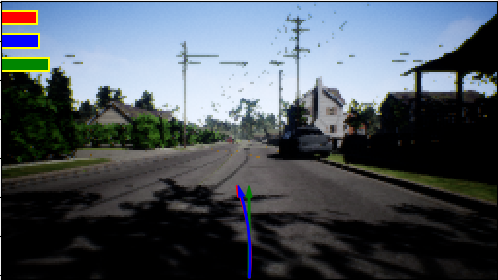}     & \includegraphics[width=0.20\linewidth]{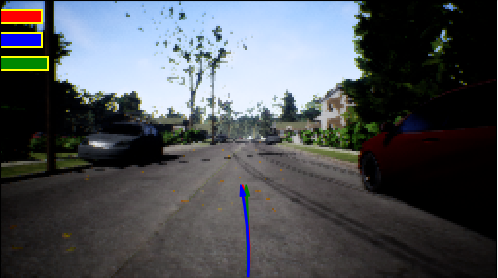} \\
    			E-PARL:0.02, E-IL:0.05 & E-PARL:0.01, E-IL:0.02 & E-PARL:0.01, E-IL:0.03 & E-PARL:0.00, E-IL:0.01 \\
    			\midrule
    			\rowcolor[rgb]{ .851,  .851,  .851} \multicolumn{4}{c}{Go straight} \\
    			\midrule
    			\includegraphics[width=0.20\linewidth]{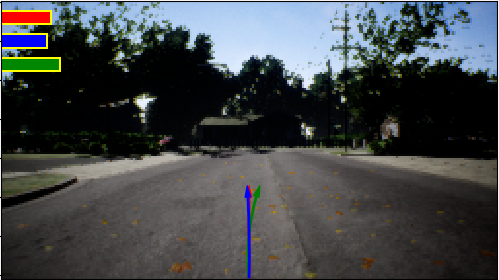}     & \includegraphics[width=0.20\linewidth]{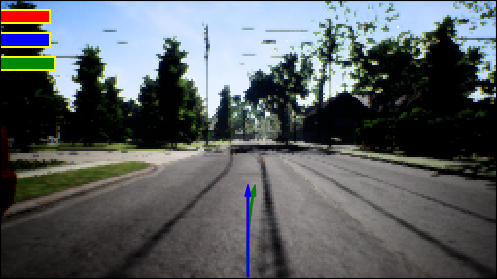}     & \includegraphics[width=0.20\linewidth]{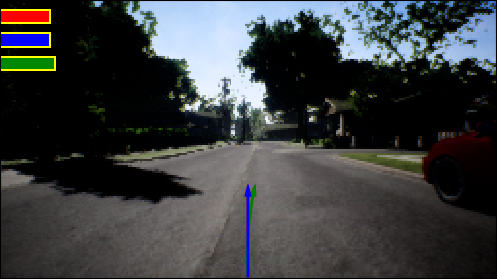}     & \includegraphics[width=0.20\linewidth]{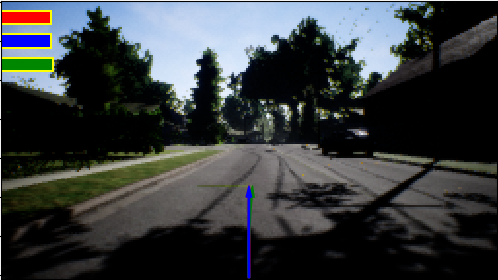} \\
    			E-PARL:0.01, E-IL:0.03 & E-PARL:0.00, E-IL:0.02 & E-PARL:0.00, E-IL:0.02 & E-PARL:0.00, E-IL:0.01 \\
    			\bottomrule
    		\end{tabular}%
    		\label{tab:addlabel}%
    		\begin{tablenotes}    
    			\footnotesize              
    			\item[1]In each figure in the table, there are three arrows and three length bars. Red represents label, blue represents PARL's output, and green represents IL's output. The direction of the arrow represents the output torque. The length of the length bar reflects the output torque value. Among them, when the torque is 0.5, it means to go straight, when the torque is less than 0.5, it means to turn left, and when the torque is greater than 0.5, it means to turn right. Below the figures are the corresponding specific values, E-PARL means errors of PARL and E-IL means errors of IL.
    		\end{tablenotes}
    	\end{threeparttable}
    \end{table*}%
    - The random resized crop method. It is used to crop and scale the image, including scale jittering method or scale and aspect ratio enhancement transformation. VGG and ResNet use this method for data augmentation.\\
    - The proposed DAT Network. Several scenario images were randomly selected and inputted to the DAT Network. Then we will obtain outputs several times the number of inputs.\\
    \textbf{Metrics.} As there are no specific quantitative metrics of data augmentation, we compare data augmentation through qualitative analysis. It includes whether the augmented data is logical, whether there is semantic augmentation, whether there is instance augmentation, the number of augmentations, etc. These factors are shown in the first line of Table III.\\
    \begin{table}[htbp]
    	\centering
    	\caption{Qualitative analysis of data augmentation methods}
    	\begin{threeparttable}
    	\begin{tabular}{ccccc}
    		\toprule
    		Methods & Number & Semantic & Instance & Reality\\
    		\midrule
    		Random Resized Crop & \cellcolor[rgb]{ .651,  .651,  .651}A+ & D (None) & D (None) & \cellcolor[rgb]{ .949,  .949,  .949}C \\
    		Color Jitter & \cellcolor[rgb]{ .816,  .808,  .808}B & D (None) & D (None) & \cellcolor[rgb]{ .651,  .651,  .651}A+ \\
    		DAT Network & \cellcolor[rgb]{ .851,  .851,  .851}A & \cellcolor[rgb]{ .749,  .749,  .749}A- & \cellcolor[rgb]{ .651,  .651,  .651}A+ & \cellcolor[rgb]{ .851,  .851,  .851}A \\
    		\bottomrule
    	\end{tabular}%
    	\label{tab:addlabel}%
    	\begin{tablenotes}    
    		\footnotesize              
    		\item[1]From A+ to D means performance from good to bad.
    	\end{tablenotes}
    	\end{threeparttable}
    \end{table}%
    \textbf{Results.} We present the result with two subsections: the implicit process of implementation for DAT Network and results comparing different methods.
    As shown in Fig.5, two scenario images are taken as examples to demonstrate the implicit process of DAT Network. The DAT Network realizes the augment of scenario in two dimensions: semantic augment and image style augment. In semantic augment, the results show that DAT Network has successfully augmented one semantic structure two times. Semantic maps reflect the logical relationship between image content and instance. From the experimental results, we can know that the DAT Network can learn the logical connection between image contents.
    Furthermore, it can generate new samples that conform to the distribution of this logical relationship. In the next, DAT Network performs the transferring between semantics to scenarios with its own P2P model. It transformed a large number of segmentation images from different agents and the cloud into scenarios. Moreover, the generated scenarios match the data distribution of the local robot. This figure demonstrates that DAT Network can generate samples corresponding to the distribution of training samples in local robots.
    \begin{table*}[htbp]
    	\centering
    	\caption{Quantitative comparison between PARL and Centralized-IL in different environments}
    	\begin{threeparttable}
    		\begin{tabular}{cccccccccc}
    			\toprule
    			\multirow{2}[4]{*}{Environment\textbackslash{}Error type} & \multicolumn{3}{c}{Environment 1} & \multicolumn{3}{c}{Environment 2} & \multicolumn{3}{c}{Environment 3} \\
    			\cmidrule{2-10}          & \cellcolor[rgb]{ .949,  .949,  .949}Obstacles & \cellcolor[rgb]{ .949,  .949,  .949}Turns & \cellcolor[rgb]{ .949,  .949,  .949}Straight & \cellcolor[rgb]{ .867,  .922,  .969}Obstacles & \cellcolor[rgb]{ .867,  .922,  .969}Turns & \cellcolor[rgb]{ .867,  .922,  .969}Straight & \cellcolor[rgb]{ .886,  .937,  .855}Obstacles & \cellcolor[rgb]{ .886,  .937,  .855}Turns & \cellcolor[rgb]{ .886,  .937,  .855}Straight \\
    			\midrule
    			\multirow{2}[4]{*}{Imitation Learning} & \cellcolor[rgb]{ .949,  .949,  .949}30.00\% & \cellcolor[rgb]{ .949,  .949,  .949}28.57\% & \cellcolor[rgb]{ .949,  .949,  .949}16.67\% & \cellcolor[rgb]{ .867,  .922,  .969}36.67\% & \cellcolor[rgb]{ .867,  .922,  .969}11.11\% & \cellcolor[rgb]{ .867,  .922,  .969}25.00\% & \cellcolor[rgb]{ .886,  .937,  .855}0.00\% & \cellcolor[rgb]{ .886,  .937,  .855}27.27\% & \cellcolor[rgb]{ .886,  .937,  .855}10.00\% \\
    			\cmidrule{2-10}          & \multicolumn{3}{c}{Overall : 27.27\%} & \multicolumn{3}{c}{Overall : 34.04\%} & \multicolumn{3}{c}{Overall : 19.05\%} \\
    			\midrule
    			\multirow{2}[4]{*}{Peer-Assisted Learning} & \cellcolor[rgb]{ .949,  .949,  .949}5.00\% & \cellcolor[rgb]{ .949,  .949,  .949}0.00\% & \cellcolor[rgb]{ .949,  .949,  .949}0.00\% & \cellcolor[rgb]{ .867,  .922,  .969}6.67\% & \cellcolor[rgb]{ .867,  .922,  .969}0.00\% & \cellcolor[rgb]{ .867,  .922,  .969}12.50\% & \cellcolor[rgb]{ .886,  .937,  .855}0.00\% & \cellcolor[rgb]{ .886,  .937,  .855}9.09\% & \cellcolor[rgb]{ .886,  .937,  .855}0.00\% \\
    			\cmidrule{2-10}          & \multicolumn{3}{c}{Overall : 3.03\%} & \multicolumn{3}{c}{Overall : 6.38\%} & \multicolumn{3}{c}{Overall : 4.76\%} \\
    			\bottomrule
    		\end{tabular}%
    		\begin{tablenotes}
    			\footnotesize              
    			\item[1]The percentage in the table indicates the percentage of controller errors in the task. Calculating it by dividing the number of errors by the number of tasks of this type.
    		\end{tablenotes}
    	\end{threeparttable}
    	\label{tab:addlabel}%
    \end{table*}%
    \begin{table*}[htbp]
    	\newcommand{\tabincell}[2]
    	\centering
    	\caption{Performance comparison between PARL and Centralized-IL in different environments}
    	\begin{threeparttable}
    		\begin{tabular}{ccccccccccccccccccc}
    			\toprule
    			& \multicolumn{18}{c}{Some Samples of result. Overall error: Centralized:0.105, PARL:0.015} \\
    			\midrule
    			\multirow{2}[4]{*}{Experiment 1} & \multicolumn{3}{c}{\includegraphics[width=0.12\linewidth]{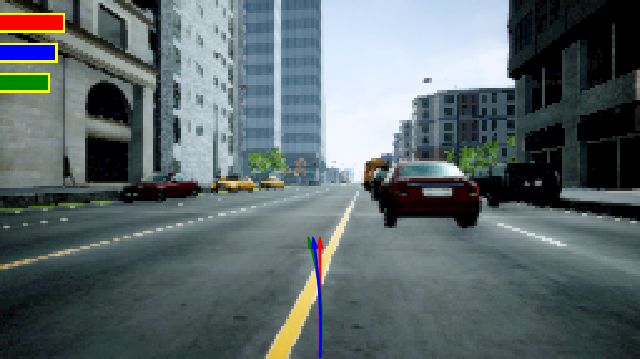}} & \multicolumn{3}{c}{\includegraphics[width=0.12\linewidth]{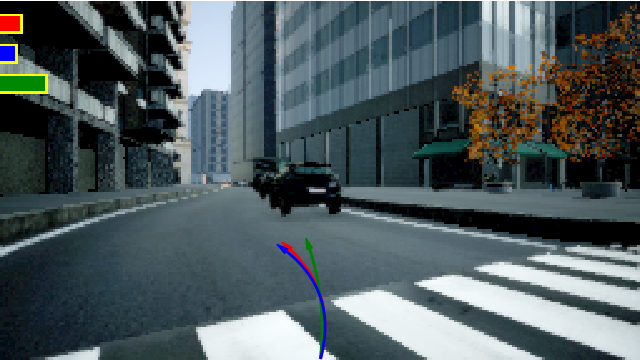}} & \multicolumn{3}{c}{\includegraphics[width=0.12\linewidth]{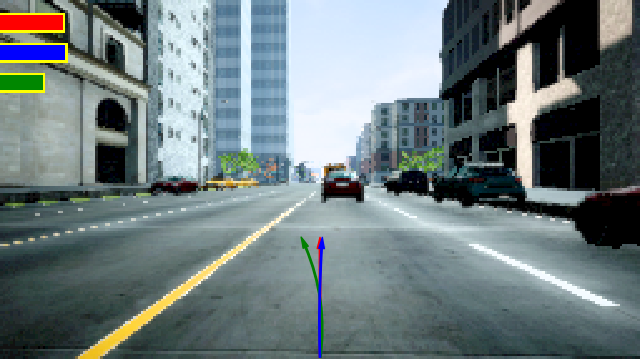}} & \multicolumn{3}{c}{\includegraphics[width=0.12\linewidth]{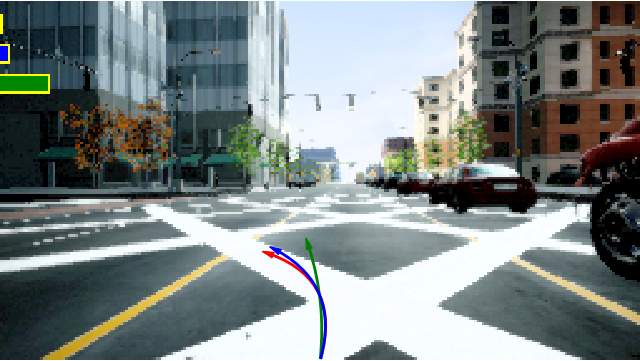}} & \multicolumn{3}{c}{\includegraphics[width=0.12\linewidth]{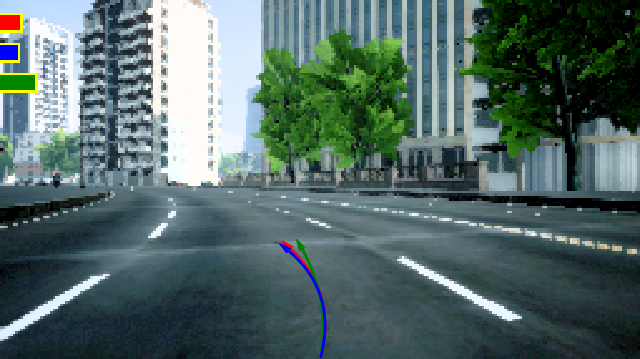}} & \multicolumn{3}{c}{\includegraphics[width=0.12\linewidth]{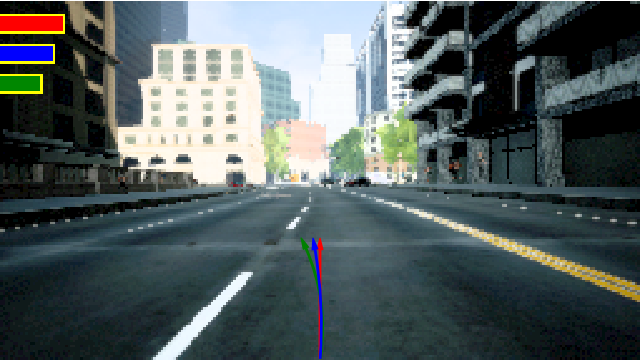}} \\
    			\cmidrule{2-19}          &  \multicolumn{3}{c}{0.49; 0.47; 0.46} & \multicolumn{3}{c}{0.31; 0.30; 0.41} & \multicolumn{3}{c}{0.50; 0.51; 0.42} & \multicolumn{3}{c}{0.29; 0.30; 0.42} & \multicolumn{3}{c}{0.37; 0.38; 0.40} & \multicolumn{3}{c}{0.50; 0.48; 0.46} \\
    			\midrule
    			\multirow{2}[4]{*}{Environment 2} & \multicolumn{3}{c}{\includegraphics[width=0.12\linewidth]{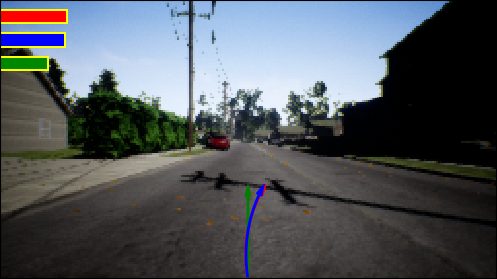}} & \multicolumn{3}{c}{\includegraphics[width=0.12\linewidth]{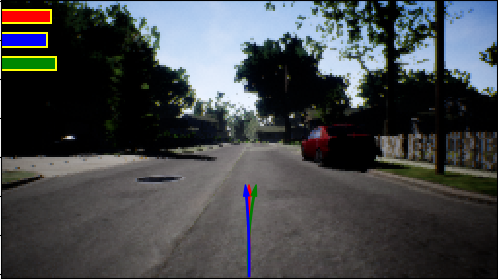}} & \multicolumn{3}{c}{\includegraphics[width=0.12\linewidth]{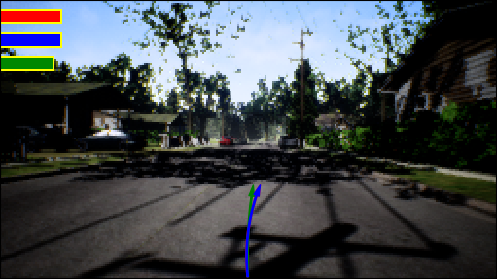}} & \multicolumn{3}{c}{\includegraphics[width=0.12\linewidth]{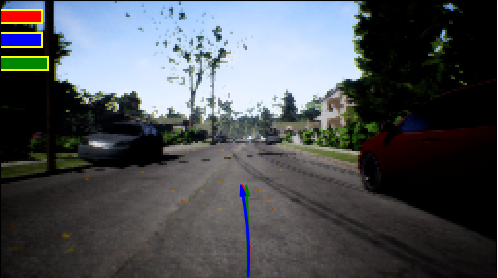}} & \multicolumn{3}{c}{\includegraphics[width=0.12\linewidth]{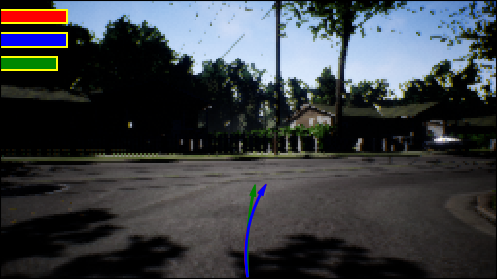}} & \multicolumn{3}{c}{\includegraphics[width=0.12\linewidth]{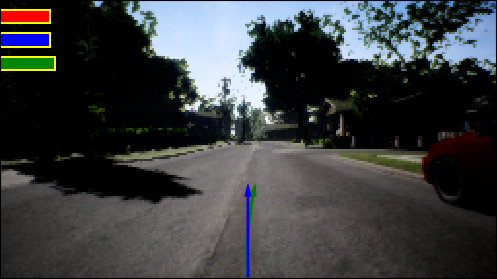}} \\
    			\cmidrule{2-19}          & \multicolumn{3}{c}{0.56; 0.55; 0.50} & \multicolumn{3}{c}{0.50; 0.49; 0.52} & \multicolumn{3}{c}{0.53; 0.53; 0.51} & \multicolumn{3}{c}{0.49; 0.49; 0.50} & \multicolumn{3}{c}{0.62; 0.62; 0.56} &
    			\multicolumn{3}{c}{0.50; 0.50; 0.51} \\
    			\midrule
    			\multirow{2}[4]{*}{Environment 3} & \multicolumn{3}{c}{\includegraphics[width=0.12\linewidth]{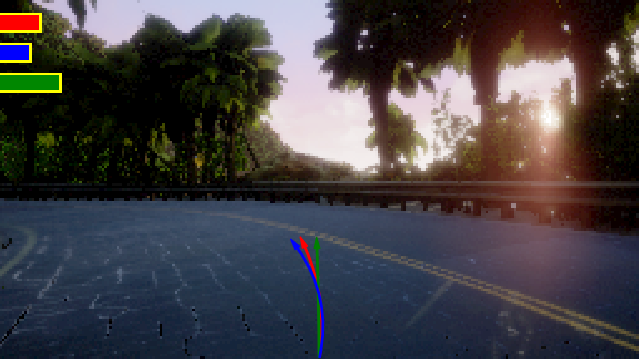}} & \multicolumn{3}{c}{\includegraphics[width=0.12\linewidth]{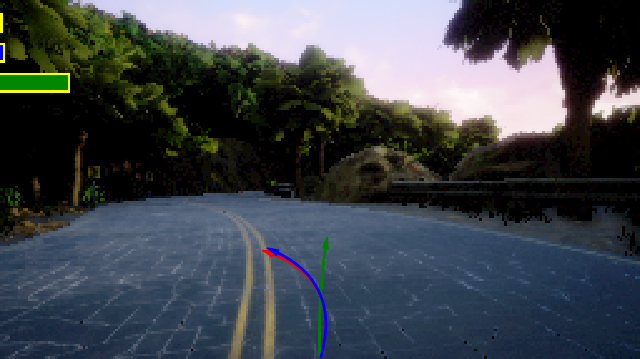}} & \multicolumn{3}{c}{\includegraphics[width=0.12\linewidth]{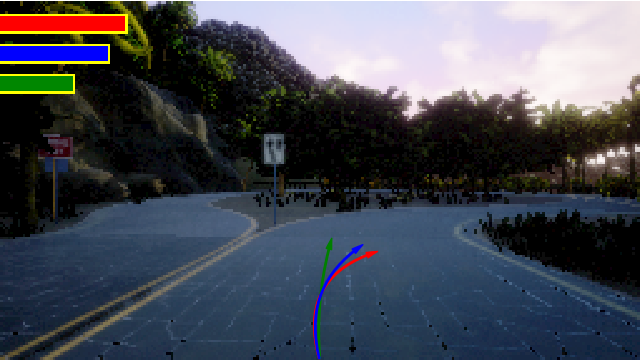}} & \multicolumn{3}{c}{\includegraphics[width=0.12\linewidth]{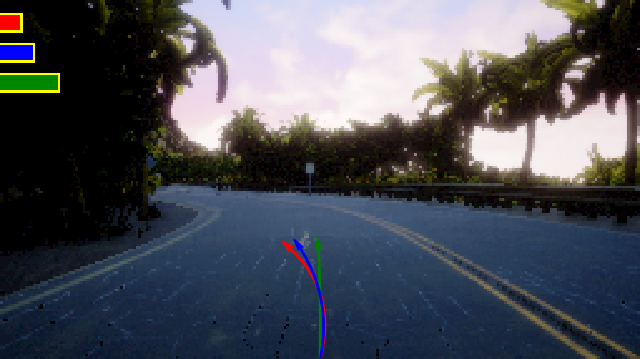}} & \multicolumn{3}{c}{\includegraphics[width=0.12\linewidth]{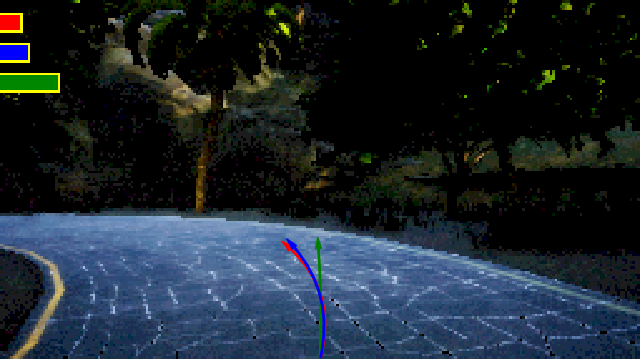}} & \multicolumn{3}{c}{\includegraphics[width=0.12\linewidth]{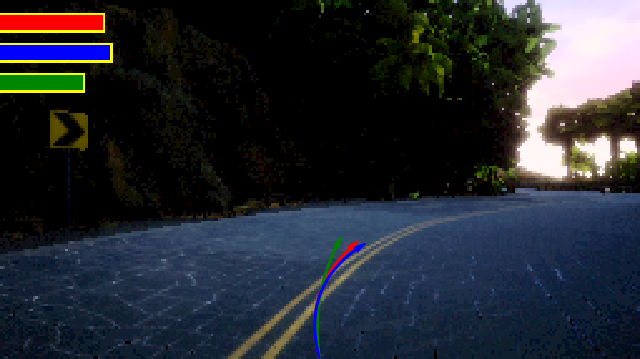}} \\
    			\cmidrule{2-19}          &  \multicolumn{3}{c}{0.44; 0.43; 0.49} & \multicolumn{3}{c}{0.30; 0.31; 0.51} & \multicolumn{3}{c}{0.71; 0.69; 0.52} & \multicolumn{3}{c}{0.40; 0.41; 0.49} & \multicolumn{3}{c}{0.39; 0.40; 0.49} &\multicolumn{3}{c}{0.62; 0.64; 0.55} \\
    			\bottomrule
    		\end{tabular}%
    		\begin{tablenotes}    
    			\footnotesize              
    			\item[1]Arrows and length bars are same to Table II. Below the figures are the corresponding specific values. Torque value of label,  Torque value of PARL output and torque value of centralized-IL are from left to right.
    		\end{tablenotes}
    	\end{threeparttable}
    	\label{tab:addlabel}%
    \end{table*}

	Augmented results comparison are presented in Table I and the qualitative analysis are shown in Table III. It can be seen from the two tables that DAT Network performed all well in "whether the augmented data is logical", "whether there is semantic augmentation", "whether there is instance augmentation", "the number of augmentations". Other methods only did well in individual indicators and did not achieve actual augmentation. Only the proposed DAT Network has achieved an increment of information in robotic learning. The availability of the augmented samples from DAT Network is higher because of the reasonable logic and style matching.

\subsection{Experiment of self-driving with PARL}
In this part, we also carried out two sub-experiments, one is to verify the effect of PARL, the other is to compare with other approaches.\\
\textbf{Experimental setup.} In work, we have used Microsoft AirSim as our simulator to evaluate the presented approach. In addition to have high-quality environments with realistic vehicle physics, AirSim has a python API that allows for accessible data collection and control. In order to collect training data, a human driver is presented with a ﬁrst-person view of the environment (central camera). The simulated vehicle is controlled by the driver using the keyboard. The car should be kept at a speed around 7m/s, collisions with vehicles or pedestrians should strive to be avoided, but trafﬁc lights and stop signs will not be considered. \\
\begin{figure}[thpb]
	\centering
	\includegraphics[width=1\linewidth]{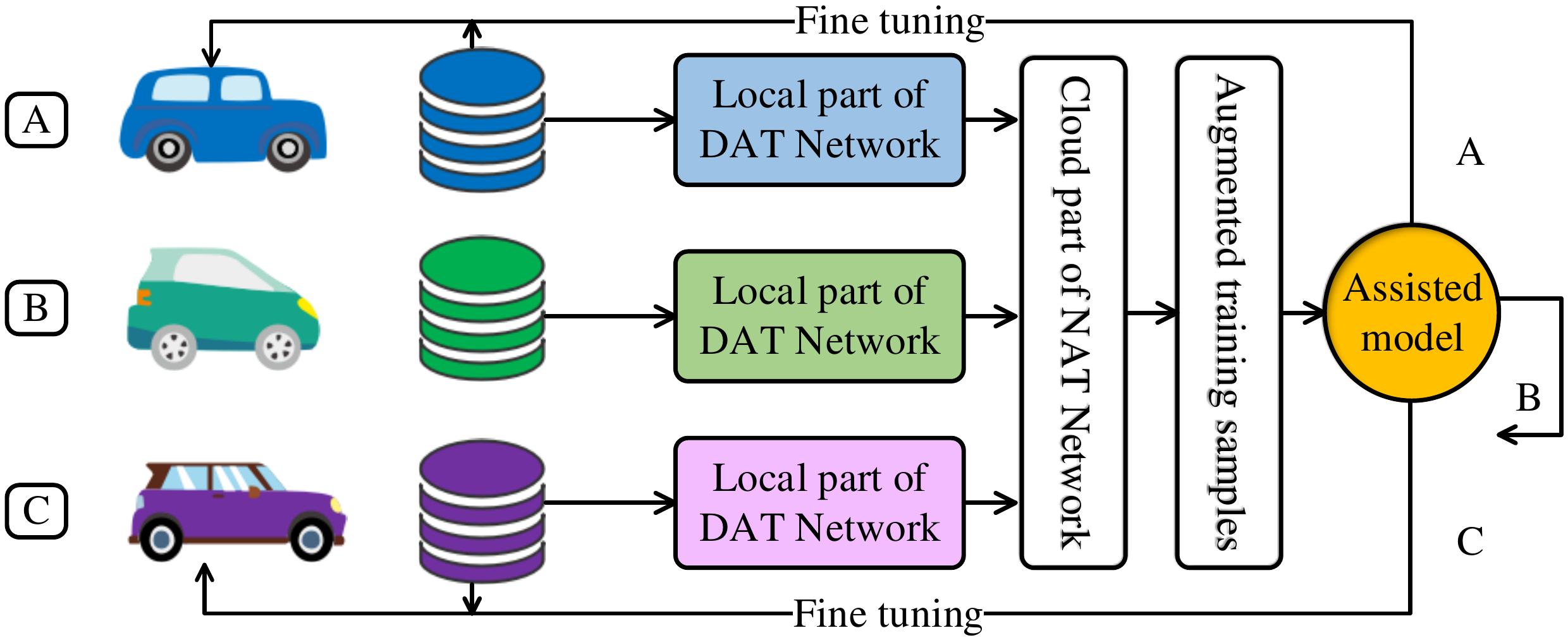}
	\caption{Experimental process of PARL. Each robot augments its dataset through DAT Network. After the cloud gets the assisted model, it is returned to the local for fine tuning. The local side computes the local part of the DAT Network and fine tunes assisted model. The cloud side computes the cloud part of DAT Network and generates labels etc.}
	\label{fig:architecture}
\end{figure}\\
\textbf{Baselines.}
The former verified experiment:\\
- Generic imitation learning (IL). The local robot performs generic imitation learning with its dataset and executes the trained policy model in the simulated environment.\\
- The proposed Peer-Assisted Robotic Learning (PARL). The local robot performs PARL in cloud robotic system and executes the trained policy model in the same environment.\\
The later comparative experiment:\\
- Policy model based on centralized datasets. Three datasets are gathered together and then performed imitation learning directly. Then, the obtained policy model is executed in the simulation environments of three agents.\\
- Policy model based on PARL. Performing self-driving learning with PARL and then execute the three models output from PARL in these three different environments.
The experimental process of PARL is presented in Fig.6. With the DAT Network, every local robot utilizes information from others to get an assisted model and fine tune it to match its style of scenarios. Finally, the fine tuned model will be executed in the corresponding experiments.\\
\textbf{Metrics.} We collected and tagged the dataset. Every label of scenario is the torque recorded from the human-drive car in the simulator environment. We compare the torque gap between the output of the model tested and the label.\\
\textbf{Results.}\\
-The former verified experiment:\\
The performance of controllers based on the two methods in key challenging tasks is presented in TABLE II. The results are summarized in Table IV. From the results, we can see that compared with general models that trained by traditional imitation learning without information from other robots, the PARL model obtained in the cloud robotic system performs significantly better in accuracy. Moreover, PARL improves the training process of imitation learning with the help of shared knowledge. There is a pre-trained model from the cloud for transferring in local imitation learning. So there is no need for local robots to learn from scratch. The transferred policies have a lower error starting point and the error value.\\
-The later comparative experiment:\\
Table IV and Table V are the performances and the summary of PARL controllers and Centralized-IL-controllers in their respective environments. It can be seen from the two tables that PARL has a better performance. Centralized-IL-controller directly centralizes the data without considering the actual contribution of the model for training. PARL considers the image style and logic in data augmentation. The data PARL generated has a higher utilization ratio, which improves the accuracy of the model.

\section{Conclusion}
This work proposes Peer-Assisted Robotic Learning to address data insufficiency and data island of local robots. PARL can enable each local robot to utilize the data and knowledge of the entire cloud robotic system. Not only does PARL converge all the data from local robots, but also it is useful for all local robots by augmenting and transferring. This greatly improves the available training samples for local robots, thus improving the training effect. Furthermore, the work presents the DAT Network to implement data augmentation in PARL, which achieves impressive results in the self-driving case. The experimental results in the self-driving case also verify the effectiveness of PARL.
\bibliographystyle{ieeetr}
\bibliography{citations}
\end{document}